\documentclass[runningheads]{template/llncs}
\usepackage[T1]{fontenc}
\usepackage{graphicx}
\usepackage{booktabs}
\usepackage{float}

\begin{document}
\title{Multi-scale Intervention Planning based on Generative Design}
%
%

\author{Ioannis Kavouras\inst{1,2}\orcidID{0000-1111-2222-3333} \and
Ioannis Rallis\inst{1,3}\orcidID{1111-2222-3333-4444} \and
Emmanuel Sardis\inst{1,4}\orcidID{2222--3333-4444-5555} \and
Eftychios Protopapadakis\inst{7,8}\orcidID{1111-2222-3333-4444} \and
Anastasios Doulamis\inst{1,5}\orcidID{2222--3333-4444-5555} \and
Nikolaos Doulamis\inst{1,6}\orcidID{2222--3333-4444-5555}
}

\authorrunning{I. Kavouras et al.}

\institute{National Technical University of Athens, Zografou Campus 9, Iroon Polytechniou str 15772 Zografou, Greece \and
\email{ikavouras@mail.ntua.gr} \and \email{irallis@central.ntua.gr} \and \email{sardism@mail.ntua.gr} \and \email{adoulam.ntua.gr} \and \email{ndoulam.ntua.gr} \and
University of Macedonia, 156 Egnatia Street, GR-546 36 Thessaloniki, Greece \and
\email{eftprot@uom.edu.gr}
}
\maketitle              
\begin{abstract}
The scarcity of green spaces, in urban environments, consists a critical challenge. There are multiple adverse effects, impacting the health and well-being of the citizens. Small scale interventions, e.g. pocket parks, is a viable solution, but comes with multiple constraints, involving the design and implementation over a specific area. In this study, we harness the capabilities of generative AI for multi-scale intervention planning, focusing on nature based solutions. By leveraging image-to-image and image inpainting algorithms, we propose a methodology to address the green space deficit in urban areas. Focusing on two alleys in Thessaloniki, where greenery is lacking, we demonstrate the efficacy of our approach in visualizing NBS interventions. Our findings underscore the transformative potential of emerging technologies in shaping the future of urban intervention planning processes.  


\end{abstract}
\section{Introduction}

The Urban green space (UGS) availability has long been investigated, because of the importance of green spaces for the health and well-being of urban residents. Generally, there are beneficial associations between green space exposure and reduced stress, positive mood, less depressive symptoms, better emotional well-being, improved mental health and behaviour, and decreased psychological distress in adolescents \cite{zhang2020association}. Yet, there is significant 
differentiation, regarding the UGS accessibility, between Northern (above-average availability) and Southern (below-average availability) European cities \cite{kabisch2016urban}. 

Generative artificial intelligence (genAI) has garnered significant attention for its transformative potential across diverse domains, including computer science, creative arts, and language processing. While its efficacy in fields like medicine and healthcare has been demonstrated, its application in engineering domains such as urban planning and architectural design remains unexplored.

In response to this gap, this paper explores the utilization of genAI, specifically generative design methodologies, in addressing critical challenges in intervention planning, particularly within urban environments. Generative design, characterized by advanced algorithms and computational techniques, offers a systematic approach to automating the generation of design scenarios based on predefined parameters and constraints. By extending its application to multi-scale intervention planning, including architectural design and urban revitalization, we aim to harness the potential of genAI in transforming urban landscapes.

The primary objective of this study is to showcase the potential of generative AI models in intervention planning applications. To this end, we introduce a simple Graphical User Interface (GUI) Desktop application developed for generating images and implementing generative design in real-world scenarios. Through experimentation and case studies, we demonstrate the feasibility and effectiveness of utilizing generative AI technology in intervention planning, thereby offering insights into its practical implications for shaping future urban environments.

The rest of this paper is organized as follow: 
(a) Section \ref{sec::RelatedWork} provides a short description of the current
literature review;
(b) Section \ref{sec::ExperimentalSetup} describes the experimental setup;
(c) Section \ref{sec::ExperimentalResults} presents the experimental results; and
(d) Section \ref{sec::Conclusions} concludes this work.

\section{Related work}
\label{sec::RelatedWork}

The generative design \cite{koenig2020integrating} for engineering applications, like urban 
planning, architectural planning, renovations and other is a relatively newly introduced field. 
Currently, this technology lacks the necessary applicability of real case experimentation
\cite{jiang2023generative}, thus further development and testing in practical and complex
design scenarios are necessary. 

In addition, it is worth mentioning that the generative design applications are not intended to replace the human factor (i.e., intelligence, opinion and design ideas). In essence, genAI models may support the architects, urban planners, etc. by proposing solutions, ideas, and scenarios. Thus, these tools can help the necessary engineers by inspiring then by the early stages of the designing process \cite{zheng2021generative}.

An indicative example can be the work of Han et al. \cite{han2021performance}, where they investigated the performance-based automatic urban design approach based on Deep Reinforcement Learning Generative Design algorithms and computer vision. By comparing their proposed approach with similar conventional approaches \cite{sun2020study,eilouti2019shape} they observed that their propose methodology is not limited by the number of the design variables, thus it can generate scenarios with different number of features, such as building of any shame. 

Another example can be the work of Zhang et al. \cite{zhang2021generative}. They developed a parametric generative algorithm for automatically generating green design scenarios of typical Chinese urban residences based on performance-oriented design flow. Similarly, Gan et al. \cite{gan2022bim} investigated the automation of the novel BIM-based graph data model based on the generative design of modular buildings. Finally, Wei et al. \cite{wei2022generative} investigate the application of the generative design approach for module construction.

The aforementioned works indicate that generative design is an emerging trend for various engineering applications and especially in architectural design, urban planning, renovation, and building construction. However, these works approach the generative design problem by developing generative algorithms similar to conventional and state-of-the-art approaches. In this work, we approach the generative design problem from a different angle. For inspiring the architects,
urban planners, and designers generally, we investigate the usage of pre-trained image-based generative models for proposing intervention ideas by the generation of photorealistic images over a predetermined area of interest. Thus, the contribution of this work is that we investigate the usage of image-to-image and image inpainting algorithms for multi-scale intervention planning. Moreover, the proposed methodology of this manuscript has not been investigated yet, thus it is important to further enrich the current literature with new methodologies based on emerging and innovative technologies.

\section{Experimental Setup}
\label{sec::ExperimentalSetup}
The main scope of this manuscript is to examine the potential of generative AI models, such as image-to-image and image inpainting techniques, for architectural design, urban planning and other forms of multi-scale intervention planning. To achieve this, we developed an application for generating images based on models from HuggingFace repository \cite{HuggingFaceModels}. The AI Image Generator Application \cite{ImageGeneratorApp} supports both image-to-image \cite{Image2ImageModel} and image inpainting \cite{InpaintingImageModel} technologies.

\begin{figure}[!htb]
  \centering
  \includegraphics[width=1.00\linewidth]{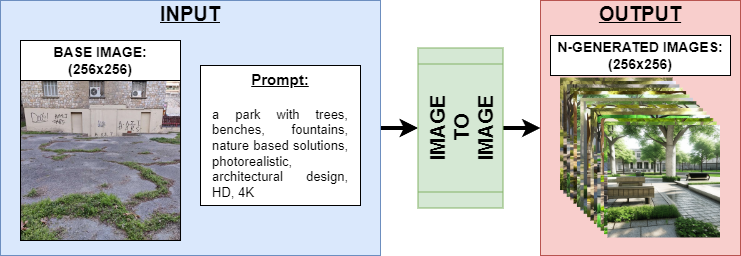}
  \caption{A brief overview of the Image to Image workflow. 
  \textbf{Input:} (a) Base Image; and (b) Description Text Prompt. 
  \textbf{Output:} N-Generated Images indicating different scenarios of intervention planning.}
  \label{fig::Image2ImageArchitecture}
\end{figure}

\begin{figure}[!htb]
  \centering
  \includegraphics[width=1.00\linewidth]{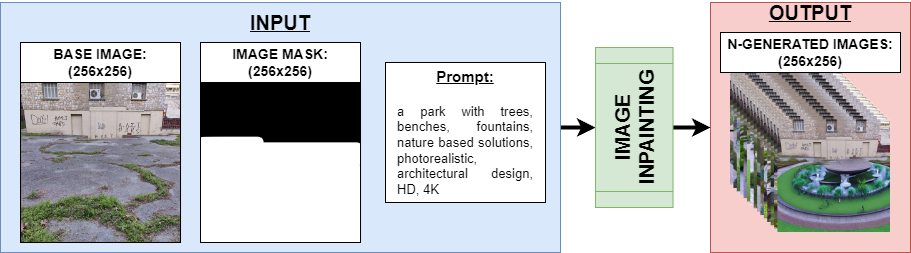}
  \caption{A brief overview of the Inpainting Image workflow. 
  \textbf{Input:} (a) Base Image; (b) Image Mask; and (c) Description Text Prompt. 
  \textbf{Output:} N-Generated Images indicating different scenarios of intervention planning.}
  \label{fig::InpaintingArchitecture}
\end{figure}

Figure \ref{fig::Image2ImageArchitecture} illustrates the image-to-image workflow, while Figure \ref{fig::InpaintingArchitecture} illustrates the image inpainting workflow. Both of these techniques need as input a base image and a description text prompt. The image inpainting needs to a mask bitmap (i.e., black and white) image for specifying the generated area of the image. The output in both techniques is a number N generated images.

\begin{figure}[!htb]
  \centering
  \includegraphics[width=0.75\linewidth]{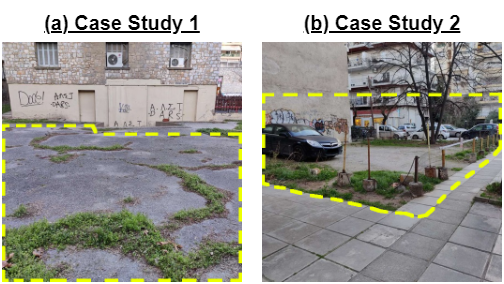}
  \caption{The images used for the experiments. The area secluded by the yellow dashed line corresponds to
  the masked area for the image inpainting technique.}
  \label{fig::CaseStudies}
\end{figure}

The proposed methodology has been evaluated over two randomly selected alleys in Thessaloniki city. These areas share some common traits as the lack of green space and the existence of high buildings in the vicinity. Thus, these two case studies can be used effectively for testing image-to-image and image inpainting techniques potential intervention planning. Figure \ref{fig::CaseStudies} illustrates the input base images. The secluded by the yellow dashed line areas indicate the inpainting masked selection, thus the area which will be changed after the potentially intervention.

The experimental results are evaluated by an architect based on the architectural composition and the realism of the generated images. Moreover, for evaluating the productivity scale of the proposed methodology and the equivalent architects work, the average time per generated image is calculated. Thus, for the experimental setup will be produced 15 images per technique per case study. In total, the final result is compromised by 60 (30 images per case study) generated images, illustrating different scenarios or possible ideas for intervene to the area of interest.

\section{Experimental Results}
\label{sec::ExperimentalResults}

\begin{figure}[!p]
  \centering
  \includegraphics[width=1.00\linewidth]{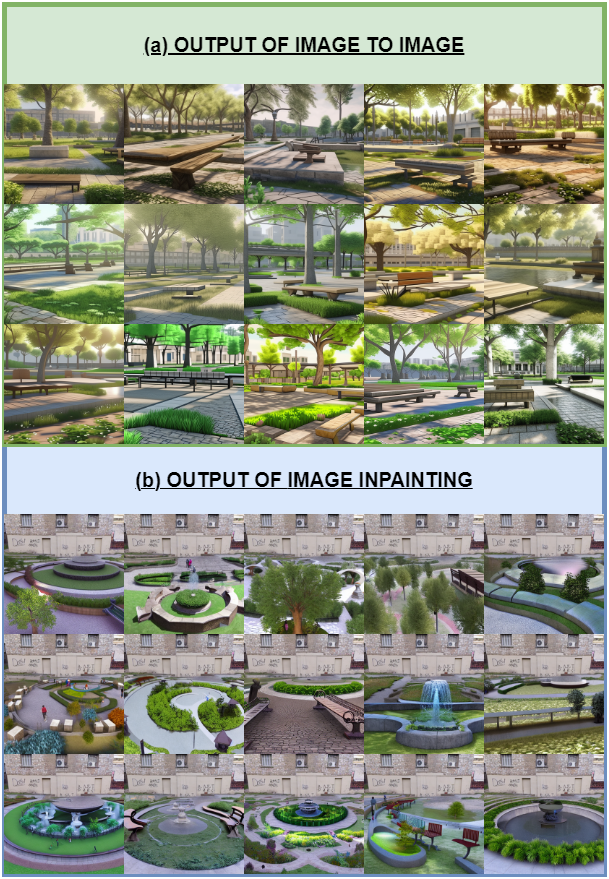}
  \caption{The generated results for the Case Study 1. (a) Image-to-Image results; and (b) Image Inpainting results}
  \label{fig::CaseStudy1}
\end{figure}

\begin{figure}[!p]
  \centering
  \includegraphics[width=1.00\linewidth]{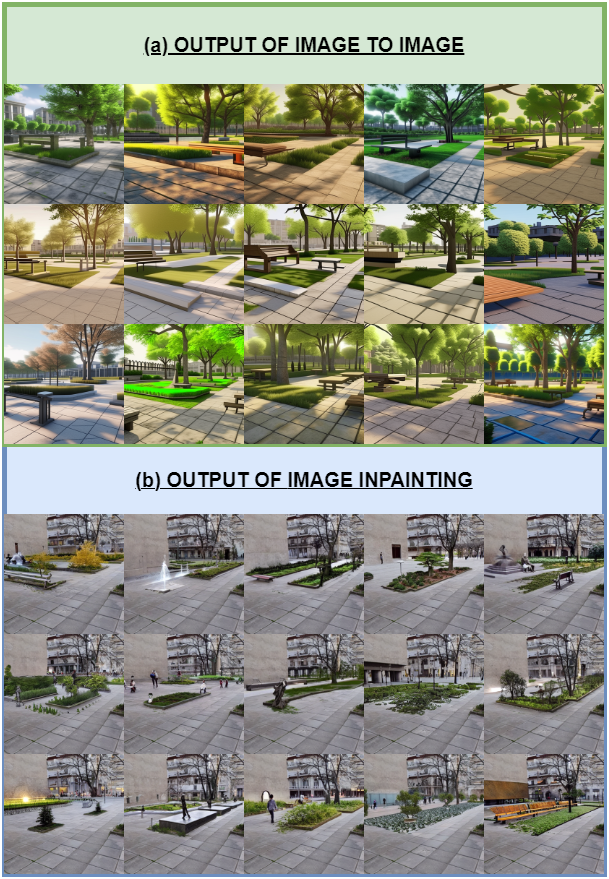}
  \caption{The generated results for the Case Study 2. (a) Image-to-Image results; and (b) Image Inpainting results}
  \label{fig::CaseStudy2}
\end{figure}

In the Section \ref{sec::ExperimentalSetup} were presented the experimental setup, which is used for generating several images of photo-realistic alternative interventions over the two case studies (Figure \ref{fig::CaseStudies}).
Figures \ref{fig::CaseStudy1} and \ref{fig::CaseStudy2} present the generated images. For each case study were generated 15 images from image-to-image method and 15 images from image inpainting method.

For the evaluation of the results we asked the opinion of architect about the architectural composition and the 
implement-ability of each solution. For the image-to-image method the generative images illustrate good looking
results, with correct photo-realism and similar to the description prompt in both case studies. However, 
image-to-image approach alters the environment of the original image and the scale of the intervention, thus the
proposed solution are not implementable as is. However, the results can inspire the architect and help him during
his architectural composition by providing him different visual ideas.

For the image inpainting approach the generative images alters only the area of intervention, thus the generative
solutions in this method are referred to the original area of interest. In many cases, the generative solution
respected the scale of the intervention as well, providing realistic and implementable results. However, the
majority of the solutions lacked the architectural composition, which means that some of the generative solutions 
are implementable, but for achieving the best possible result, the architect needs to further process the solution.

In addition, based on architects' design methodology, the related needed time to produce a photo-realistic similar to the generative images varies from 1 to 4 hours, depended on the architectural composition, the scale of the area and the depth of details, as well as the software that the photo-realistic will be produced. For the generative methods, the image-to-image approach needed approximately 3 minutes to produce a 256x256 generative image, using CUDA acceleration and an NVidia 3070 Ti 8GB graphic card. Using the same setting, the image inpainting approach needed approximately 4 minutes to produce an image. Table \ref{tab::TimeCompare} summarizes the time differences for generating a photo-realistic images for intervention planning.

\begin{table}[!htb]
    \centering
    \caption{Time duration for generating a photorealistic image of pixel size 256x256.}\label{tab::TimeCompare}
    \begin{tabular}{p{4cm} c c}
        \toprule
        \textbf{Generating Method} & \textbf{Time} & \textbf{Images per Hour}\\
        \midrule
        Architect & 1-4 hours & max 1\\
        \hline
        Image-to-Image & $\sim$3 minutes & $\sim$20\\
        \hline
        Image Inpainting & $\sim$4 minutes & $\sim$15\\
        \bottomrule
    \end{tabular}
\end{table}

In general, the generative AI technology can really help the experts (i.e., architects, urban planners and other 
engineers in the design field). Even if the generated solutions are not the best possible, compared with and 
architect's composition, they still can be used for inspiring the experts. A strong advantage of this technology is
that they can produce several solutions in a short time. Thus, this technology can be sufficiently used for intervention
planning, however further research is necessary.

\section{Conclusions}
\label{sec::Conclusions}

In conclusion, this study delves into the potential of generative AI for multi-scale intervention planning, with a focus on addressing urban green space scarcity. The experimentation in Thessaloniki's alleys, Greece, employing image-to-image and image inpainting methods, highlights genAI's promising role in architectural and urban planning. These models swiftly propose diverse solutions, providing valuable assistance even in preliminary planning phases. Despite challenges posed by current models trained on generalized data, specialized datasets and ongoing research promise to enhance their implementation. By overcoming these obstacles, generative AI can revolutionize intervention planning, fostering sustainable and vibrant urban landscapes.

\begin{credits}
\subsubsection{\ackname}
This work is supported by the European Union Funded project euPOLIS ”Integrated NBS-based Urban Planning Methodology for Enhancing the Health and Well-being of Citizens: the euPOLIS Approach”, under the Horizon 2020 program H2020-EU.3.5.2., grant agreement No. 869448.
\end{credits}
%
%
%
\bibliographystyle{splncs04}
\bibliography{mybibliography}
%
\end{document}